# COMPARISON BETWEEN GENETIC FUZZY METHODOLOGY AND Q-LEARNING FOR COLLABORATIVE CONTROL DESIGN

Anoop Sathyan, Kelly Cohen and Ou Ma

Department of Aerospace Engineering, University of Cincinnati, Cincinnati, OH

*ABSTRACT*

*A comparison between two machine learning approaches viz., Genetic Fuzzy Methodology and Q-learning, is presented in this paper. The approaches are used to model controllers for a set of collaborative robots that need to work together to bring an object to a target position. The robots are fixed and are attached to the object through elastic cables. A major constraint considered in this problem is that the robots cannot communicate with each other. This means that at any instant, each robot has no motion or control information of the other robots and it can only pull or release its cable based only on the motion states of the object. This decentralized control problem provides a good example to test the capabilities and restrictions of these two machine learning approaches. The system is first trained using a set of training scenarios and then applied to an extensive test set to check the generalization achieved by each method.*

## 1. INTRODUCTION

This paper discusses a comparison between two Machine Learning (ML) methodologies, viz. Genetic Fuzzy Methodology (GFM) and Q-learning, to design controllers for a set of collaborative robots that should work together to achieve a common goal without the need for any explicit inter-robot communication. There has been a lot of research conducted in the field of collaborative robotics. These include development of controlled physical compliance for external contacts [1-5] that could be useful for human-robot collaborative tasks, swarm intelligence control algorithms [6-9] as well as multi-robot collaboration with minimal communication between the robots [10, 11].Such intelligent collaborative robots can help in various applications such as material handling [12], mapping the interior of buildings [13], exploration [14], factory automation [15] etc., to name a few.

This research focuses on a different type of problem where a team of independently controlled robots work together to achieve a common goal while they are also physically connected to an object through elastic cables. The collaboration enables the total workload on the system to be shared among the set of robots. Such a decentralized system is applicable to various collaborative applications. Specifically in the field of robotics, such a team of decentralized controllers can be used for (a) lifting or moving tasks that involve multiple collaborative robots or human-robot collaboration [5,12], (b) robotic soccer where team of robots have to work together to achieve the common objective of scoring more goals than the opponents [16,17], (c) swarm of Unmanned Aerial Vehicles (UAVs) [18] that work together on reconnaissance missions, just to name a few. The advantage of developing decentralized controllers is that the success of the team is not just dependent on one centralized controller. In centralized control applications, if the centralized controller malfunctions, then the entire system fails, whereas when using a series of decentralized controllers, even if one of the individual controllers were to fail, the rest of the system may still be able to achieve the overall goal. We can also say that as the size of the team of robots increase, the dependency on a single individual decreases.

For the problem considered in this paper, the robots are trained to work together using two separate ML methodologies, GFM and Q-learning, for comparing the two methodologies with each other. Since there is no communication between the robots,e ach robot is unaware of the state and specific future action of the partner robots although all the robots are aware of their common goal.

The last decade has seen a huge rise in the use of machine learning approaches, mainly due to the increase in computational capability as well as accessibility to huge amounts of data. As these intelligent systems learn from data, it provides adaptability, scalability, robustness to uncertainties etc.Another advantage of intelligent systems is that it providesthe ability to make decisions based on a variety of inputswhich in turn leads to increased efficiency.

Fuzzy logic system (FLS) is one such intelligent system. As fuzzy logic provides a smooth transition between the fuzzy sets, FLSs provide an inherent robustness to the design of robotic controllers. Although expert knowledge canbe used to build FLSs and this capability is appealing to alot of applications, it makes sense to have a mechanism to tune the parameters of the FLS automatically using a search heuristic such as Genetic Algorithm (GA). This methodology of using GA to train an FLS is called a GFM and the resulting system is known as a Genetic Fuzzy System (GFS). Such GFSs have been developed with much success for clustering and task planning [19], simulated air-to-air combat [20], aircraft conflict resolution [21] etc. An FLS design requires a set of membership functions for each input and output variable, as well as a rule base for designing the relationship between the input and output variables. Since it is trained using GA, differentiable cost function such as integral squared error is not required. So, as long as the mission requirement can be defined using a mathematical cost function, we do not need to have any ground truth data available. GA will traverse the search space looking for the optimal set of membership functions and rule base that minimizes the cost function, which makes it a form of reinforcement learning.

This paper presents a comparison of the GFM with Q-learning, which is widely regarded as the current state-of-the-art in the field of reinforcement learning. Q-learning approach involves creating a dataset of states and actions which is then used to train an Artificial Neural Network (ANN) that outputs the best action based on the current input state. Control agents can be trained using reinforcement learning to take optimal actions at every instant to reach a final desired state. Q-learning, which is a form of reinforcement learning, has gained a lot of popularity recently in training ANNs and Convolutional Neural Networks (CNNs) for various applications including training agents to autonomously play Atari games [22], the development of the Alpha Go system that defeated professional human Go players [23] etc.

Our previous works [24, 25] showed the effectiveness of the GFM to three and five robot collaborative systems. In this paper, we build upon those previous efforts to do a comparison study of our GFSs with those trained using Q-learning. In the GFM, GA is used to tune the parameters of the FLS. In the Q-learning approach, Q-learning algorithm is used to create a dataset of states and corresponding optimal actions that is then used to train an ANN. Both GFS and ANN can model nonlinear systems very well. GFS has the added advantage of being inherently robust, although ANNs can also be trained to achieve improved robustness.

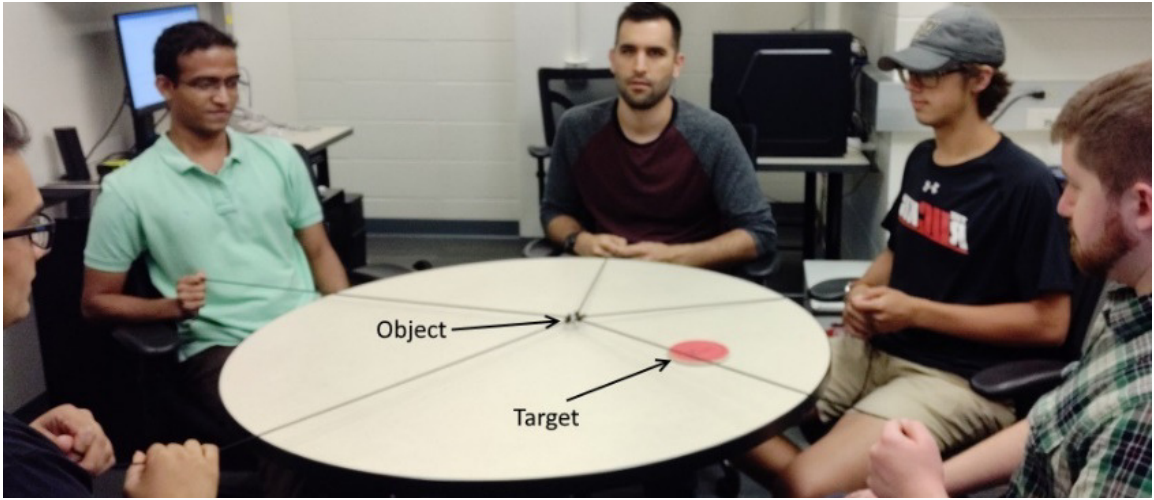

Figure1. The research work presented in this paper takes inspiration from this game played collaboratively by a group of people. The objective of the individuals is to collaboratively control the cables to bring the object to the target position [25].

## 2. PROBLEM DESCRIPTION

This problem is inspired from a game involving people working collaboratively to bring an object to a target position by pulling or releasing the cables, as shown in Figure 1. The participants do not communicate with each other. This game showcases human ability to learn and adapt to situations that require collaboration. As each human makes their own decisions, this game can also be considered as a decentralized control problem. Taking inspiration from this game, we are developing decentralized control strategy and algorithms to allow individual robots to perform similar activities showing the capability of multi-robot collaboration. Although humans are very adept at these kinds of collaborative activities, it isquite challenging for robots due to the current limitations ofrobot intelligence. The robots have to learn to work together in order to achieve their common goal.

The motion plane of the robots and the object is assumed to be horizontal. The robots are fixed at the vertices of a regular polygon. The robots can only pull or release the cable attached to it. A top-down view of the setup for the5-robot case is shown in Figure 2. The robots are placed at a distance of 0.5m from the center. The objective is to have the robots to work collaboratively to bring the object to an arbitrarily defined target position by pulling or releasing the elastic cables that are connected to the object. One major constraint is that each robot only has information about the target and the object and does not have any knowledge about the states of the partner robots. Thus, this problem provides a great example to test the capability of robots to work collaboratively without the need for any centralized control or inter-robot communication. The robots need to be trained for different scenarios to come up with an effective strategy to achieve the common goal while following all the constraints of the problem. The training is done using GFMas well as using Q-learning in order to perform a comparison between the two approaches.

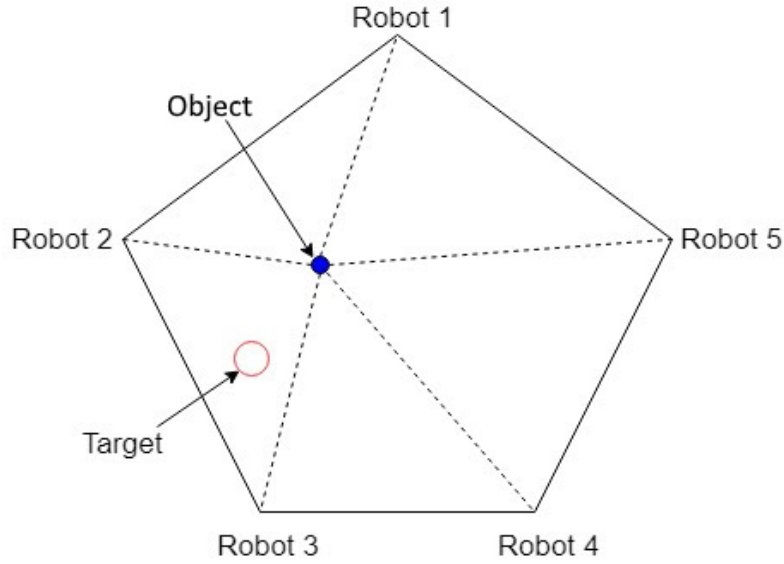

Figure 2. Setup for the 5-robot problem. The robots are placed on the vertices of a regular pentagon. The dotted lines are the cables connecting the object to each of the robots.

## 3. SYSTEM DYNAMICS

The equations of motion for an N-robot system is givenbelow [24, 25].

$$k \sum_{i=1}^{N}(p_i - l_0)\widehat{r_{iB}} - N r_B = m \ddot{r}_B \qquad (1)$$

Eqn. (1) is a 2-D vector equation pertaining to the motion of the object which is connected to the robots through the elastic cables. The vectors in the equation can be understood from Figure 3, which shows the vector representations for a 5-robot problem. $\widehat{r_{iB}}$ is the unit vector along the line connecting the object B to robot $i$. $p_i$ refers to the length of the cable reeled in by robot $i$. Eqn. (1) is valid when all the cables are taut, i.e. the length of the cables are within 1-2m. If any of the cables go slack, the tension in that cable can be considered as zero. The maximum length of the cables is considered as 2m, beyond which the cables break.

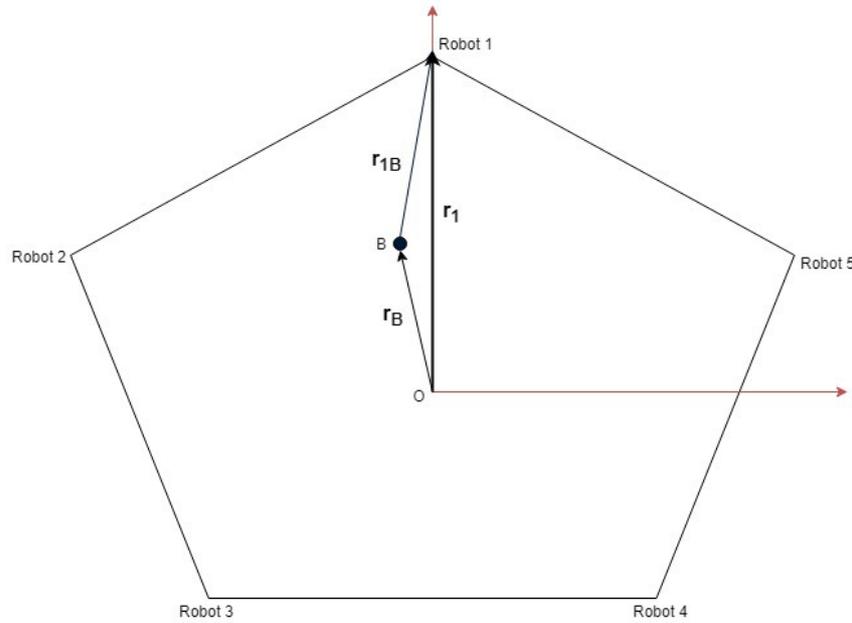

Figure 3. Relation between the position vectors for the object B and robot1, shown for the 5-robot system [25].

Eqn. (1) shows that the dynamics of the object is dependent on the lengths of the cables reeled in by each robot. Each robot can control its cable through a setup of DC motor and spool around which the cable can wind. For the sake of brevity, we do not delve into the dynamics of the DC motors that control the spools. But, it is to be noted that the controller for each robot directly controls the voltage of the motor which in turn causes the spool to rotate, providing each robot the capability to pull or release its cable to control the position of the object. The objective of this problem is to train these decentralized robots to work collaboratively to achieve the common goal of bringing the object to any predefined position within the workspace of the robots.

## 4. METHODOLOGIES

### 4.1. Genetic fuzzy methodology

Each robot is modeled as a GFS. Through the training process assisted by GA, the robots learn to work together to achieve the common goal without the need for any centralized control. The schematic of the GFS controller for robot *i* is shown in Figure 4. Each GFS takes in four inputs and gives one output. The inputs to each GFS controller include the distance between the current object position and the target position measured with respect to the vector connecting the robot to the target, and the angle between the object-robot vector and the target-robot vector. Additionally, the object velocity along the x and y axes arealso provided as inputs. The object velocity helps the robots to understand the current direction of motion of the object. We believe these four inputs should be sufficient to make a good decisions by each of the robots in order to collaboratively achieve the common goal. The GFS outputs a voltage, V, at each time-step which is used to control how much the robot pulls or releases the cable at that time-step.

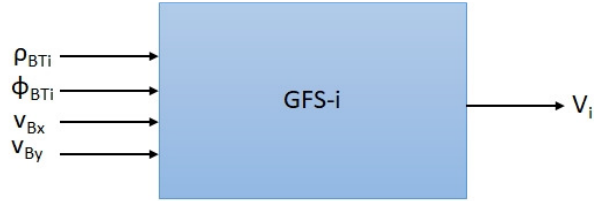

$\rho_{BTi}$ = radial distance between ball and target as measured from the robot i

$\phi_{BTi}$ = angle between vectors $r_{iT}$ and $r_{iB}$

$v_{Bx}$ = current velocity of the ball along the x-axis

$v_{By}$ = current velocity of the ball along the y-axis

Figure 4. Schematic of the GFS controller for robot *i*

The robots are trained on a set of training scenarios. The target positions are arbitrarily defined for these 20 scenarios. During training, each individual in GA, which consists of a vector of GFS parameters, is evaluated on the training set of scenarios. The object is assumed to always start at the origin. During the training process, GA tunes the membership functions and rulebase for each controller to minimize the mean of the following cost function that is evaluated for each scenario.

$$C = \int_0^T \text{dist}(t)dt + 50(T - t_{end}) \qquad (2)$$

T is the maximum time and $t_{end}$ refers to the time at which the simulation stops. The simulation stops when maximum time, T, is reached or when any cable length becomes greater than 2m or in other words, any of the cable breaks. In Eqn. (2), the *50(T-$t_{end}$)* term is used to penalize such early stoppage situations, where the multi-robot system is not satisfying the physical constraints. dist(t) is the distance between the object and the target at each time step. Thus, the objective is to bring the object to the target position within a minimum time, while following the constraints of the system.

The schematic of the training process is shown in Figure 5.GA starts off with a set of individuals for the population. Each individual is a vector that consists of parameters for all the robots in the system. These parameters include the membership function boundaries as well as the consequents of the rule base. For each individual in GA, the scenario could be simulated to evaluate the cost function defined by Eqn. (2). The individuals with lower cost values have more likelihood of being selected for crossover and mutation and being chosen into the next generation. The individualswith high cost values have a greater likelihood of getting kicked out of the population. This process of modifying the individuals through crossover and mutation continues for a predefined number of generations. During each generation, the best system of robots is also evaluated on a validation set. The validation set consists of new scenarios with arbitrarily defined target positions different from those in the training set. The validation allows us to check if the team of robots are generalizing well on new scenarios. After GA has reached he maximum number of generations, the individual with the best training and validation cost is chosen. This individual defines the trained system of collaborative robots that can work together to achieve the common goal.

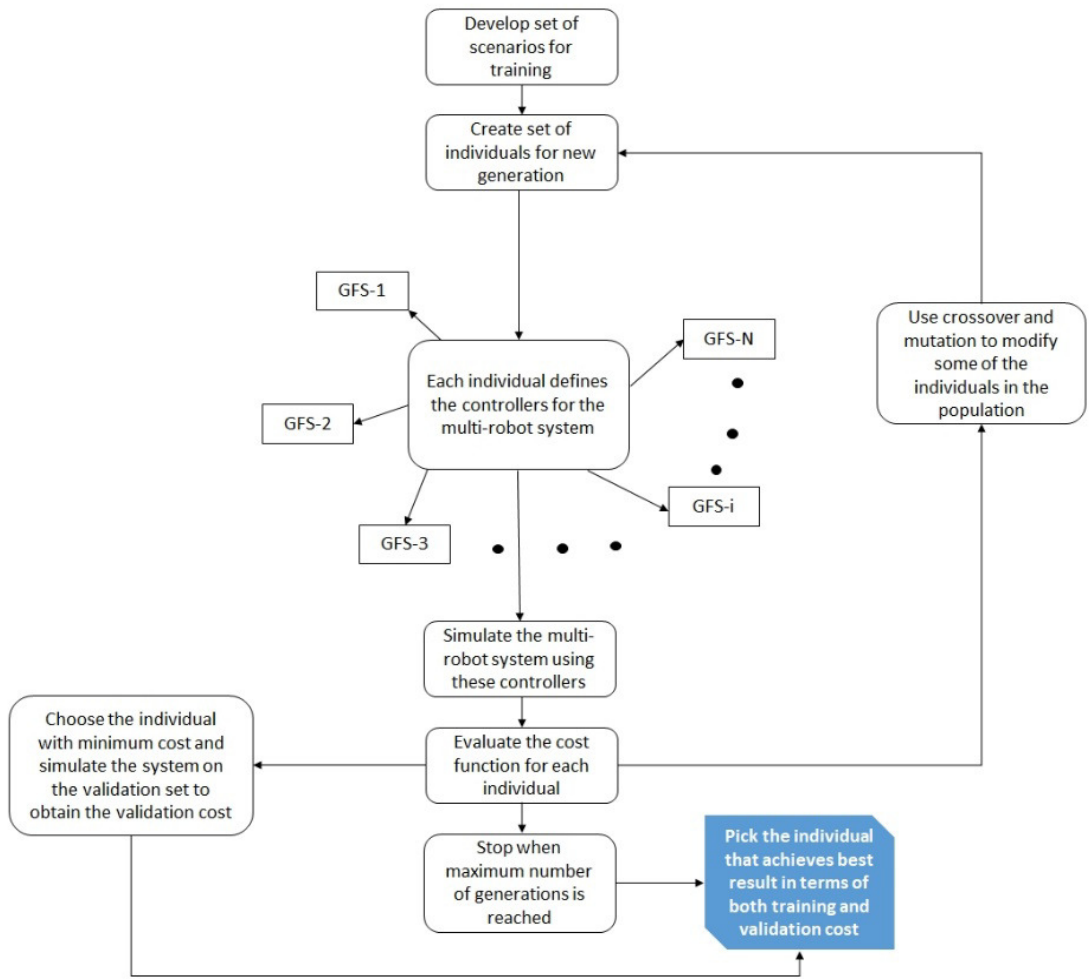

Figure 5. Training process for the N-robot system. The parameters of each robot are tuned using GA simultaneously to minimize the cost function.

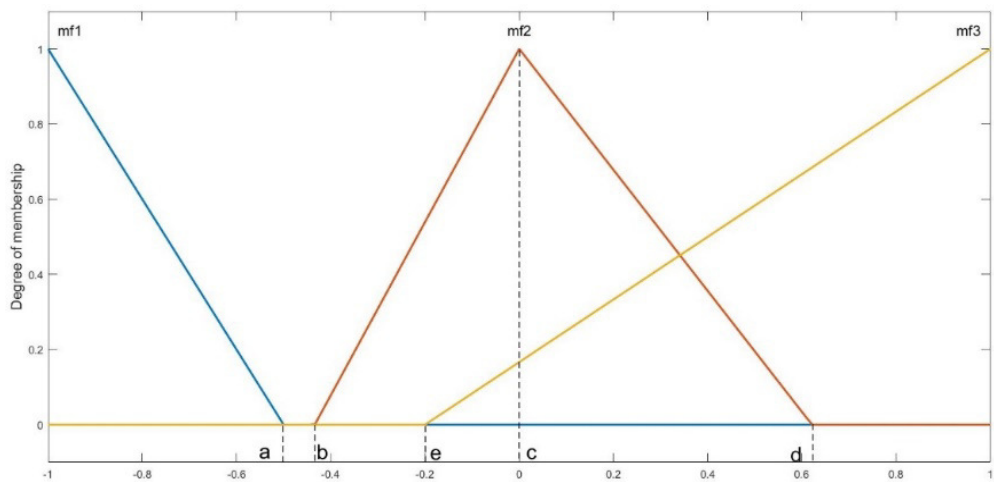

Figure 6. Membership function: The 5 points a, b, c, d, e are tuned for each input using GA.

Figure 6 shows the structure of the membership functions for each input variable. Three triangular membership functions are used for each input. The y-axis of Figure 6 shows the degree of membership of a value of particular input variable. The membership values of each input variable is needed when evaluating the linguistic rules in the rule base [24].In order to do that, the membership functions of each input variable need to be defined. As three triangular membership functions are used for each variable, the boundaries of the three triangles need to be defined. It is assumed that the membership functions, *mf1 and mf3*, peak at the left and right extremes, respectively. The x-coordinates of the five vertices are tuned using GA for each input variable. The number of membership functions for each input and output variable is chosen such that it provides the robots enough capability to learn as well as generalize. Having larger number of membership functions will increase the learning capacity, but also increases chances of overfitting. As can be seen from Figure 6, the membership functions are modeled as triangles and GA tunes five membership function parameters for each input variable. GA tunes only one boundary of each of the two extreme membership functions viz., *mf1* and *mf3*, while tuning all three parameters of *mf2*.

On the other hand, the output variable is defined using five triangular membership functions and all vertices of the five triangles are tuned using GA. This means that GA tunes 15 parameters of the output membership functions for each robot. This provides sufficient learning capacity for the robots. Additionally, GA also tunes the rule base of the GFS for all the robots. Since each GFS has four inputs and each input variable is defined using three membership functions, there will be *$3^4$ = 81* rules in the rule base of each GFS. The number of membership functions can be modified according to the problem, if needed. Some of the rules can be predefined which can reduce the search space for GA. In this work, no such assumptions are made and GA is used to tune the entire rule base of each robot.

### 4.2. Q-learning

We also apply the Q-learning methodology to train each robot. This would require developing a Q-table, where each row consists of a state-action pair along with the corresponding Q-value by running various training scenarios. The Q-value is a measure of the quality of the action $a_t$ taken at the current state, $s_t$. For ourN-robot problem, we will have N separate Q-tables. During the training process, we start off with empty Q-tables and as we run the N-robot system for various scenarios, each Q-table gets populated with the state-action pairs and their corresponding Q-values. At any instant, to be consistent with our GFS schematic, each robot has 4 states and one output, viz. the voltage, applied to the controller (*V*) that creates the pulling or releasing action on the cable. Thus, the Q-table for each robot will have six columns. At each time step, each of the *N*robots perform *N* actions that moves the object to anew position. Based on the object movement, we can assign a reward to each robot (at time *t*) that is evaluated according to Eqn. (3). Here, $r_{BT}$ represents the connecting object B to target T. As seen from Eqn. (3), the robots obtain a positive reward for every time the object is moved closer to the target.

$$R_t = \|\mathbf{r_{BT}}(t)\| - \|\mathbf{r_{BT}}(t+1)\| \tag{3}$$

Thus, the robots collect rewards at each time step toachieve the overall goal of bringing the object to the targetposition. After each time step, the Q value corresponding tostate action pair ($s_t$, $a_t$) is updated for robot *i* using Eqn. (4).α is the learning rate whereas γ is called the discount factor,which is a measure of the importance of future rewards ascompared to the current reward.

$$Q_i^{new}(s_t, a_t) = (1 - \alpha)Q_i(s_t, a_t) + \alpha(R_t + \gamma . \max_a(Q(s_{t+1}, a))) \tag{4}$$

Figure 7 shows the schematic of the development process of the Q-tables for the team of robots. As mentioned before, each robot has a Q-table associated with it during training. The Q-tables get populatedwith different state-action pairs and their correspondingQ-values as the system of robots encounter new states during the training runs. After simulating the system over several training scenarios, the system will achieve a reduced cost value, defined in Eqn. (2). At this stage, we can use the Q-tables to create a dataset of states and best actions for each robot. The best action for any state is the one that has the highest Q-value. This is done for all $N$ robots and these $N$ different datasets can then be used to train ANNs. This has the added advantage that we are not limited to discrete states and actions. The ANN will provide an action output for any continuous state. Thus, each robot is modeled as an ANN that takes in the four state values and provides an estimate of the best action ($V_i$). The ANN used for this work has only one hidden layer with 30 neurons, as this provided good fitting and generalization on the Q-table. As the data is obtained from running the robots in the collaborative environment, the $N$ different robots should be able to work together to achieve their desired common goal. The trained system of robots can then be tested on a large test set.

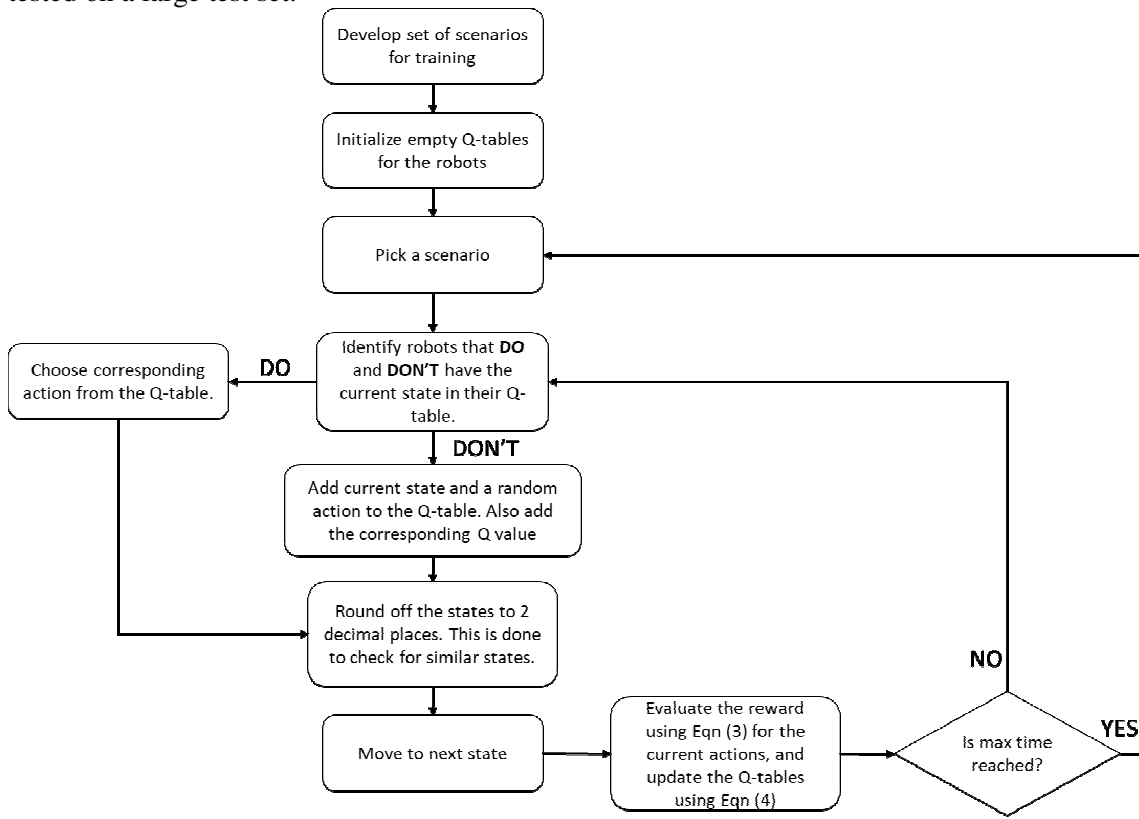

Figure 7 - Schematic that shows the development of the Q-tables for the multi-robot system.

## 5. RESULTS

Both GFS and Q-learning approaches were applied to the problem consisting of three and five robots. The different training scenarios are created using different target locations. It makes sense to have a diverse set of scenarios that have the target positions scattered over different regions of the control space of the robots. This ensures that the team of robots, after training, is able to bring the object to almost any target location within their control space. Once the systems are trained separately using GFM and Q-learning, the multi-robot systems trained using both methods were tested on 100 different scenarios for both the three robot as well as the five robot cases.Their performance on one of the three robot and five robot scenarios are shown in Figures8and 9, respectively. The main observations and comparisons are as follows:

1) As can be seen from Figures 8 and 9, the GFS controllers were able to settle the object to the target position faster. But, it should be noticed that the ANNs provide a smoother path as compared to the GFS controllers. The reason for the faster settling time could be because of the cost function in Eqn. (2) used to train the GFS controllers that is evaluated over the entire scenario as opposed to the rewards in Eqn. (3) (for Q-learning) that just considers the reward based on the current time step.

2) Both GFS and ANN were tested on 100 different scenarios for both the 3-robot and 5-robot cases. Both GFS and ANN achieved the final goal in all of the 100 scenarios tested in the case of 3-robot problem. For the 5-robot problem, the GFS was able to bring the object to the target position for 88 scenarios whereas ANNs trained using Q-learning did the same for only 82 scenarios. It is possible that this performance could be improved with more training of both the systems.

3) It was noticed that GFM required more training time as compared to Q-learning, especially in the case of the 5-robot problem. For the 5-robot case, GFM needed 50% more training time than Q-learning. This was expected as the cost function (in Eqn. (2) used for evaluating the GFS requires the entire scenario to be simulated and this needs to be done for each individual in GA over a number of generations.

4) Both GFS and ANN required more number of training scenarios and more training time for the 5-robot case as compared to the 3-robot cases. Since we are dealing with collaborative robots that do not communicate their states or actions with each other, the problem gets more complicated as robots are added to the system. On the other hand, having more number of robots increases the workspace of the system and the reliability of success as the success of the team is not dependent on a single robot.

5) Finally, it was also noticed that the GFS controllers required a much smaller training set as compared to ANN to achieve generalization. For example, the GFS controllers for the 5-robot problem were trained on 12 scenarios whereas the ANN controllers required 50 different training scenarios.

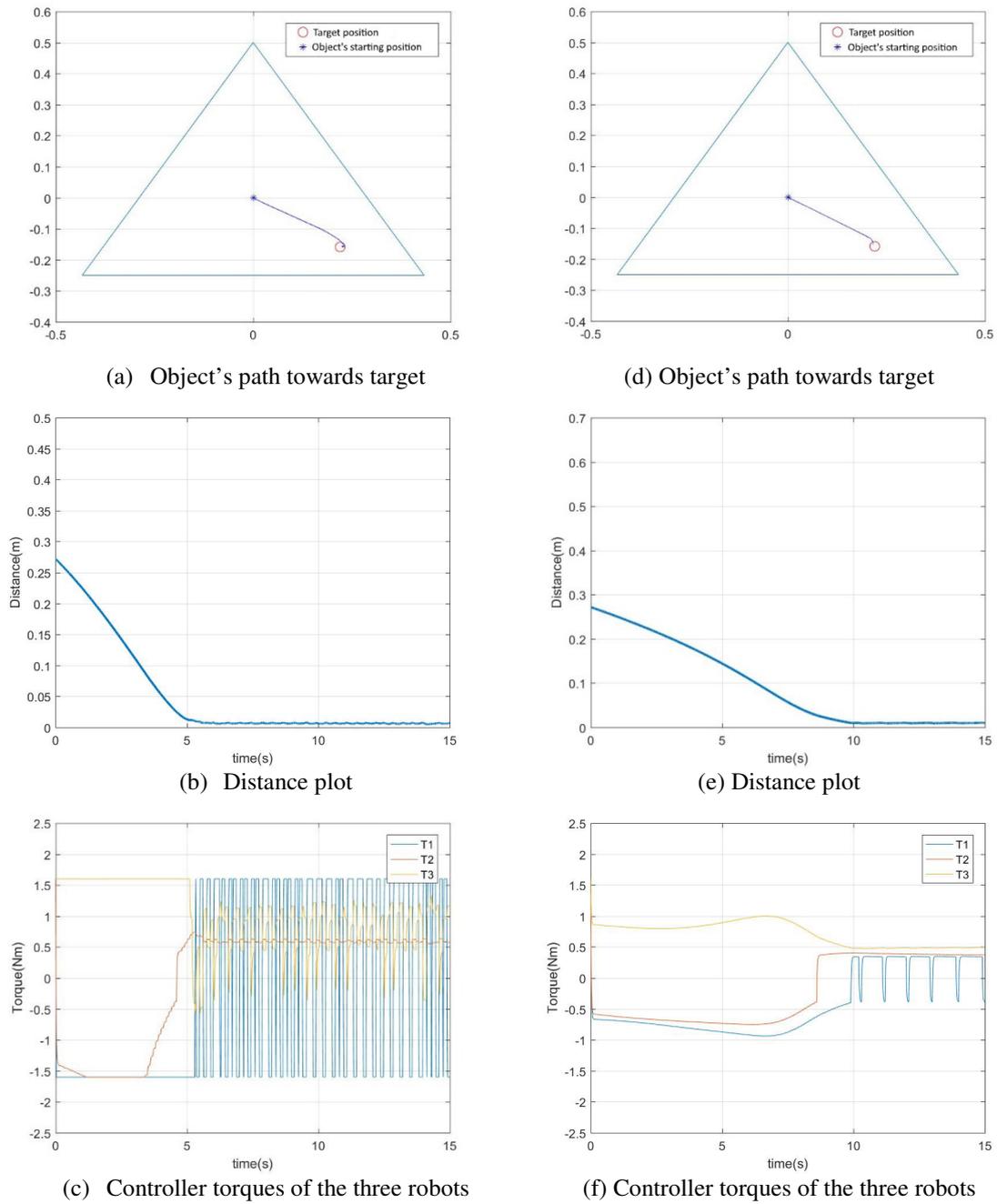

Figure 8. GFS v/s Q-learning: A 3-robot scenario. The left column (a)-(c) shows the results obtained using GFS controllers, and the right column (d)-(f) shows the results for the same scenario obtained using ANN controllers trained using Q-learning.

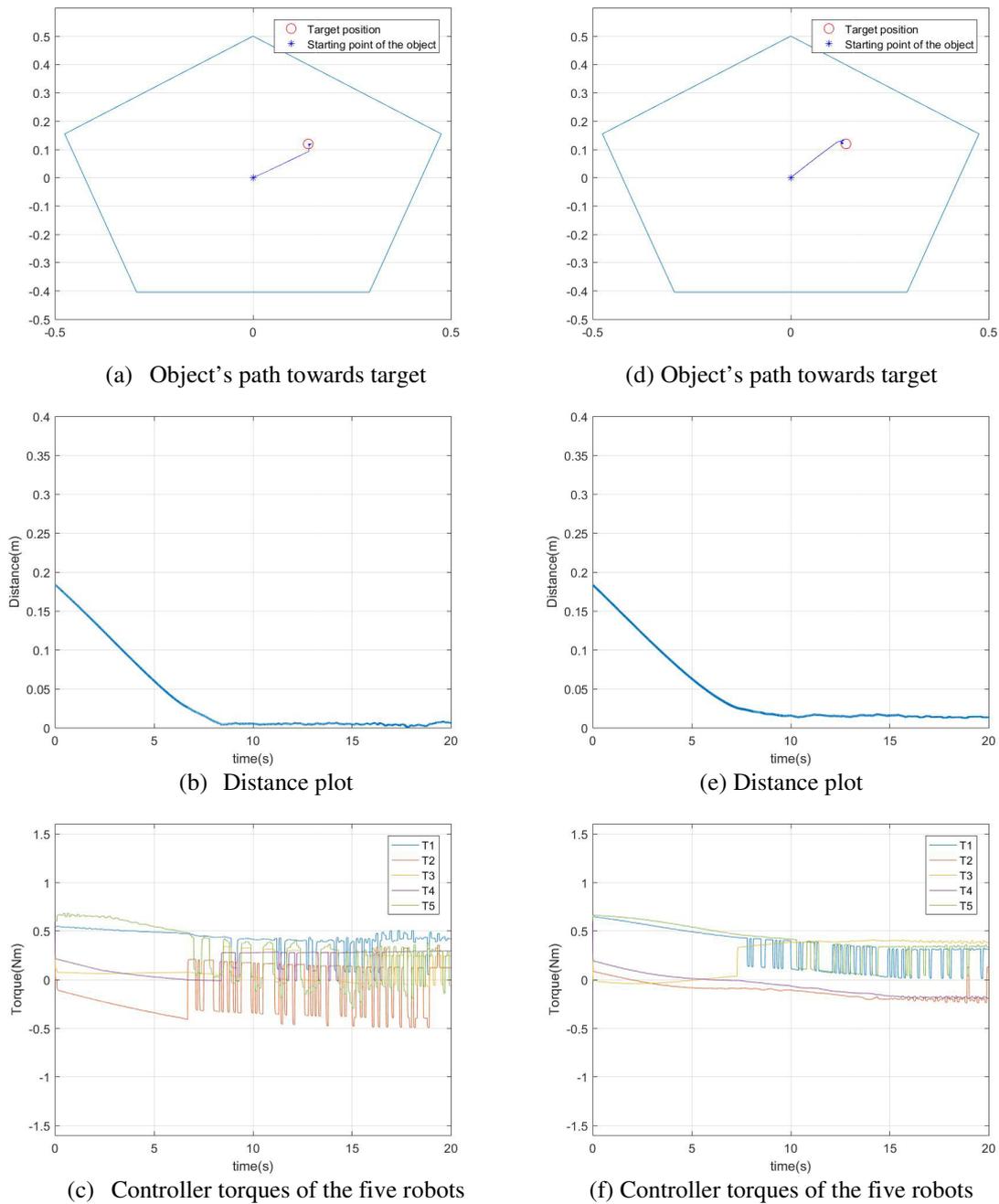

Figure 9. GFS v/s Q-learning: A 5-robot scenario. The left column (a)-(c) shows the results obtained using GFS controllers, and the right column (d)-(f) shows theresults for the same scenario obtained using ANN controllers trained using Q-learning.

## 6. CONCLUSIONS

A comparative study between GFM and Q-learning for a class of multi-robot collaborative control problem has been presented. The problem discussed had an additional constraint that the robots cannot communicate with each other. Thus, the methods presented had to train the individual robots to work collaboratively to achieve the common goal of bringing the object to any arbitrarily defined target position without any inter-robot communication. By applying GFM and Q-learning separately for solving this problem for the case involving three and five robots, we were able to do a comparative study and show the pros and cons of the two approaches. The GFM required a smaller training set of scenarios compared to Q-learning, but the Q-learning methodology provides faster training. It was also seen that the success rate of the GFM was higher than that of Q-learning for the case of the five robots. We also proved that these machine learning approaches are scalable even though the problem becomes more complex as more robots are added .In the future, we plan to expand on this work for larger number of robots. Future work will involve testing the scalability of these approaches to systems consisting of larger number of robots.

The use of decentralized control and the lack of any inter-robot communication reduces any overhead requirement. Such a decentralized methodology also ensures that even if one of the robots were to malfunction, the system will still be able to function even though the overall functionality maybe reduced. This will be especially true for problems with larger number of robots.

The trained system of robots were able to bring the object to the target region very quickly (less than 20s) using both GFM and Q-learning. The team of robots performed so efficiently in spite of several constraints considered in the system including the maximum length of the cable, the limited degrees of freedom of the robots, etc.

**REFERENCES**

[1] L. Rozo, S. Calinon, D. G. Caldwell, P. Jimenez, and C. Torras, "Learning physical collaborative robot behaviors from human demonstrations," IEEE Transactions on Robotics, vol. 32, no. 3, pp. 513–527,2016.

[2] L. Wang, "Collaborative robot monitoring and control for enhanced sustainability." International Journal of Advanced Manufacturing Technology, vol. 81, 2015.

[3] A. M. Djuric, R. Urbanic, and J. Rickli, "A framework for collaborative robot (cobot) integration in advanced manufacturing systems," SAE International Journal of Materials and Manufacturing, vol. 9, no. 2,pp. 457–464, 2016.

[4] N. L. Tagliamonte, F. Sergi, D. Accoto, G. Carpino, andE. Guglielmelli, "Double actuation architectures for rendering variableimpedance in compliant robots: A review," Mechatronics, vol. 22,no. 8, pp. 1187–1203, 2012.

[5] M. Geravand, F. Flacco, and A. De Luca, "Human-robot physicalinteraction and collaboration using an industrial robot with a closedcontrol architecture," in Robotics and Automation (ICRA), 2013 IEEEInternational Conference on. IEEE, 2013, pp. 4000–4007.

[6] M. Brambilla, E. Ferrante, M. Birattari, and M. Dorigo, "Swarmrobotics: a review from the swarm engineering perspective," SwarmIntelligence, vol. 7, no. 1, pp. 1–41, 2013.

[7] M. A. Joordens and M. Jamshidi, "Consensus control for a system ofunderwater swarm robots," IEEE Systems Journal, vol. 4, no. 1, pp.65–73, 2010.

[8] M. A. Ma'sum, G. Jati, M. K. Arrofi, A. Wibowo, P. Mursanto,and W. Jatmiko, "Autonomous quadcopter swarm robots for objectlocalization and tracking," in Micro-NanoMechatronics and HumanScience (MHS), 2013 International Symposium on. IEEE, 2013, pp.1–6.

[9] H. Duan and P. Qiao, "Pigeon-inspired optimization: a new swarmintelligence optimizer for air robot path planning," InternationalJournal of Intelligent Computing and Cybernetics, vol. 7, no. 1, pp.24–37, 2014.

[10] T. A. Phan and R. A. Russell, "A swarm robot methodology forcollaborative manipulation of non-identical objects," The InternationalJournal of Robotics Research, vol. 31, no. 1, pp. 101–122, 2012.

[11] Z. Wang and M. Schwager, "Multi-robot manipulation without communication,"in Distributed Autonomous Robotic Systems. Springer,2016, pp. 135–149.

[12] E. Gambao, M. Hernando, and D. Surdilovic, "A new generation ofcollaborative robots for material handling," in ISARC. Proceedings ofthe International Symposium on Automation and Robotics in Construction,vol. 29. Vilnius Gediminas Technical University, Departmentof Construction Economics & Property, 2012, p. 1.

[13] N. Michael, S. Shen, K. Mohta, V. Kumar, K. Nagatani, Y. Okada,S. Kiribayashi, K. Otake, K. Yoshida, K. Ohno, et al., "Collaborativemapping of an earthquake damaged building via ground and aerialrobots," in Field and Service Robotics. Springer, 2014, pp. 33–47.

[14] Senthilkumar, K.S. and Bharadwaj, K.K., 2011. "Player-stage based simulator for simultaneous multi-robot exploration and terrain coverage problem". International Journal of Artificial Intelligence & Applications, 2(4), p.123.

[15] J. A. Marvel, "Collaborative robots: A gateway into factory automation," Retrieved from thomasnet.com, 2018.

[16] Mendoza, J.P., Biswas, J., Cooksey, P., Wang, R., Klee, S., Zhu, D. and Veloso, M., 2016, March. Selectively reactive coordination for a team of robot soccer champions. In Thirtieth AAAI Conference on Artificial Intelligence.

[17] Fang, N.C., Tsai, T.N., Wu, L.F., Cheng, C.H., Huang, C.Y., Liu, C.Y. and Li, T.H.S., 2017, November. "Multi-robot coordination strategy for 3 vs. 3 teen-sized humanoid robot soccer game". In 2017 International Automatic Control Conference (CACS) (pp. 1-6). IEEE.

[18] Choi, J.Y. and Kim, S.G., 2012. "Collaborative tracking control of UAV-UGV". World Academy of Science, Engineering and Technology, 71.

[19] A. Sathyan, N. D. Ernest, and K. Cohen, "An efficient genetic fuzzyapproach to UAV swarm routing," Unmanned Systems, vol. 4, no. 02,pp. 117–127, 2016.

[20] N. Ernest, D. Carroll, C. Schumacher, M. Clark, K. Cohen, and G. Lee,"Genetic fuzzy based artificial intelligence for unmanned combat aerialvehicle control in simulated air combat missions," J Def Manag, vol. 6,no. 144, pp. 2167–0374, 2016.

[21] A. Sathyan, N. Ernest, L. Lavigne, F. Cazaurang, M. Kumar, andK. Cohen, "A genetic fuzzy logic based approach to solving theaircraft conflict resolution problem," in AIAA Information Systems-AIAA Infotech@ Aerospace, 2017, p. 1751.

[22] V. Mnih, K. Kavukcuoglu, D. Silver, A. A. Rusu, J. Veness, M. G.Bellemare, A. Graves, M. Riedmiller, A. K. Fidjeland, G. Ostrovski,et al., "Human-level control through deep reinforcement learning,"Nature, vol. 518, no. 7540, p. 529, 2015.

[23] D. Silver, A. Huang, C. J. Maddison, A. Guez, L. Sifre, G. VanDen Driessche, J. Schrittwieser, I. Antonoglou, V. Panneershelvam,M. Lanctot, et al., "Mastering the game of Go with deep neuralnetworks and tree search," Nature, vol. 529, no. 7587, pp. 484–489,2016.

[24] A. Sathyan and O. Ma, "Collaborative control of multiple robots usinggenetic fuzzy systems approach," in ASME Dynamic Systems andControl Conference, 2018.

[25] Sathyan, A., Ma, O. and Cohen, K., 2018. Intelligent Approach for Collaborative Space Robot Systems. In 2018 AIAA SPACE and Astronautics Forum and Exposition (p. 5119).

## Biography

**Anoop Sathyan** works as a Senior Research Associate in the Department of Aerospace Engineering at University of Cincinnati. Dr. Sathyan obtained his B.Tech from College of Engineering, Trivandrum, M.E from Indian Institute of Science and PhD from University of Cincinnati. His research interests include dynamics and control, genetic fuzzy systems, machine learning, neural networks and optimization. His PhD thesis focused on applying machine learning methods, mainly genetic fuzzy, for different aerospace applications. During his PhD work at UC, he has also collaborated with University of Bordeaux in France and Psibernetix Inc. for broadening the applicability of genetic fuzzy methodology. Currently, as a Senior Research Associate, he focuses on applying his knowledge in genetic fuzzy systems towards various control applications in the field of robotics and aerospace engineering.

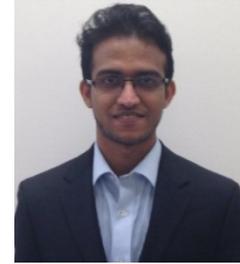

**Kelly Cohen**, the Brian H. Rowe Endowed Chair in aerospace engineering, has been a faculty member at UC's College of Engineering and Applied Science, or CEAS, for more than 10 years and currently serves as interim head of the Department of Aerospace Engineering and Engineering Mechanics. His career is marked by achievement in the field of aerospace engineering and education, including the UC Dolly Cohen Award for Excellence in Teaching, the American Institute of Aeronautics and Astronautics Outstanding Technical Contribution Application Award, the CEAS Distinguished Researcher Award and the Greater Cincinnati Consortium of Colleges and Universities Excellence in Teaching Award, among many others.

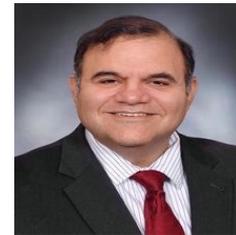

**Ou Ma** obtained his B.Sc. degree from Zhejiang University, and M.Sc. and Ph.D. degrees from McGill University. His research interests are in dynamics and intelligent control of robotic systems for space, manufacturing, and biomedical applications. Dr. Ma is currently an endowed chair professor in the Department of Aerospace Engineering and Engineering Mechanics, University of Cincinnati. He is leading an effort to develop a research program in Intelligent Robotics and Autonomous Systems (IRAS) at UC. Prior to joining UC in 2017, Dr. Ma was the Nakayama Professor at the New Mexico State University, where he developed three major research labs and accomplished many federally sponsored research projects in contact dynamics, flight dynamics and controls, and aerospace robotics. Prior to joining NMSU in 2002, Dr. Ma had been employed by MDA Space Missions (MD Robotics) for over eleven years, as a project engineer and R&D technical lead, developing modeling, simulation, and experimental verification techniques for the space robotic systems: Canadarm (SRMS), Canadarm2 (SSRMS) and Dextre (SPDM). Dr. Ma also worked as a visiting professor at German Space Center (DLR) in 2010 and Tsinghua University in 2015-2016 while on sabbatical leaves from NMSU.

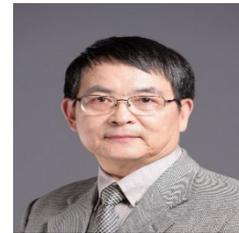